\documentclass[journal]{IEEEtran}

\usepackage{graphicx}
\usepackage{listings}

\begin{document}

\title{Integration of Large Language Models in Control of EHD Pumps for Precise Color Synthesis}

\author{Yanhong~Peng$^{1,2,*}$,
        Ceng~Zhang$^{3}$,
        Chenlong~Hu$^{4}$
        and~Zebing~Mao$^{5,*}$
\thanks{$^{1}$Department of Information and Communication Engineering, Graduate School of Engineering, Nagoya University, Nagoya 4648601, Japan.}
\thanks{$^{2}$College of Mechanical Engineering, Chongqing University of Technology, Chongqing 400054, China.}
\thanks{$^{3}$Department of Mechanical Engineering, National University of Singapore, Singapore 119077, Singapore.}
\thanks{$^{4}$Department of Computer Science and Technology, Tsinghua University, Beijing, 100084, China.}
\thanks{$^{5}$Department of Mechanical Engineering, Tokyo Institute of Technology, Tokyo 152-8550, Japan}
\thanks{$^{*}$Correspondence: Y. Peng (e-mail: yhpeng@nagoya-u.jp), Z. Mao (e-mail: mao.z.aa@m.titech.ac.jp).}
        }

\markboth{Journal of \LaTeX\ Class Files
}%
{Shell \MakeLowercase{\textit{et al.}}: Bare Demo of IEEEtran.cls for IEEE Journals}

\maketitle

\begin{abstract}
This paper presents an innovative approach to integrating Large Language Models (LLMs) with Arduino-controlled Electrohydrodynamic (EHD) pumps for precise color synthesis in automation systems. We propose a novel framework that employs fine-tuned LLMs to interpret natural language commands and convert them into specific operational instructions for EHD pump control. This approach aims to enhance user interaction with complex hardware systems, making it more intuitive and efficient. The methodology involves four key steps: fine-tuning the language model with a dataset of color specifications and corresponding Arduino code, developing a natural language processing interface, translating user inputs into executable Arduino code, and controlling EHD pumps for accurate color mixing. Conceptual experiment results, based on theoretical assumptions, indicate a high potential for accurate color synthesis, efficient language model interpretation, and reliable EHD pump operation. This research extends the application of LLMs beyond text-based tasks, demonstrating their potential in industrial automation and control systems. While highlighting the limitations and the need for real-world testing, this study opens new avenues for AI applications in physical system control and sets a foundation for future advancements in AI-driven automation technologies.
\end{abstract}



\section{Introduction}

\IEEEPARstart{I}{n} the continuously evolving fields of automation and artificial intelligence, the integration of complex Large Language Models (LLMs) with physical control systems represents a frontier area with immense potential. Currently, LLM are employed in various sectors including human-robot interaction \cite{zhang2023large}, medicine \cite{thirunavukarasu2023large}, and education \cite{chang2023survey}, benefiting from their robust text comprehension, generalization capabilities, and customizable and scalable nature. These models, trained on extensive textual data, have acquired the ability to understand and generate natural language. They are not only capable of processing data types encountered during their training but also generalize to new tasks and domains. Additionally, these models can be fine-tuned according to specific application requirements to better adapt to particular tasks or fields \cite{ding2023parameter}. Furthermore, with advancements in computational power and algorithmic optimization, the scale and capabilities of these models are continually increasing \cite{austin2021program}.

Electrohydrodynamic (EHD) pumps utilize the principles of electro-fluid dynamics to move and control fluids. These pumps, devoid of traditional rotating impellers or pistons, exhibit minimal wear and require low maintenance \cite{peng2023review}. The simple structure of EHD pumps makes them suitable for miniaturization and portable devices \cite{mao2023soft}. The motivation for this study stems from the growing demand for automation in fields requiring precise color mixing, such as printing, textile manufacturing, and digital art creation. Although traditional color mixing methods are effective, they often lack the precision and adaptability needed for complex or large-scale applications. The emergence of EHD pump technology has opened new avenues for precise fluid manipulation. However, the development efficiency of EHD control is currently low due to the complexity of programming and control, and the full potential of these pumps has not been fully realized.

This paper proposes a framework for controlling EHD pumps using LLMs, employing fine-tuned models to interpret natural language commands and convert them into precise operational instructions for Arduino-controlled EHD pumps. It explores the potential for innovative applications at the intersection of computational linguistics, robotics, and material science. The primary objective of this framework is to achieve accurate and customizable color synthesis by manipulating the flow of primary color liquids (yellow, magenta, and cyan) through these pumps. Additionally, by fine-tuning LLMs, the framework offers high scalability, allowing the free addition of various functional modules to achieve customized tasks. Utilizing the capabilities of advanced language models, particularly those developed by OpenAI (such as GPT-4), this research aims to bridge the gap between user-friendly interfaces and complex hardware control. The language model acts as an intermediary, translating user-defined color specifications expressed in natural language into specific executable Arduino code. This approach not only simplifies the interaction between users and hardware but also enhances the precision and flexibility of the color synthesis process.

The main contribution of this paper is to demonstrate the framework of combining cutting-edge artificial intelligence technology with physical control systems, paving the way for more intuitive, efficient, and multifunctional automation solutions in various industrial and creative fields. The originality of this research lies in its pioneering approach to directly integrating LLMs like GPT-4 with industrial control systems, specifically in the context of operating EHD pumps for precise color synthesis.

\begin{figure*}[!t]
\begin{center}
\includegraphics[width=12cm]{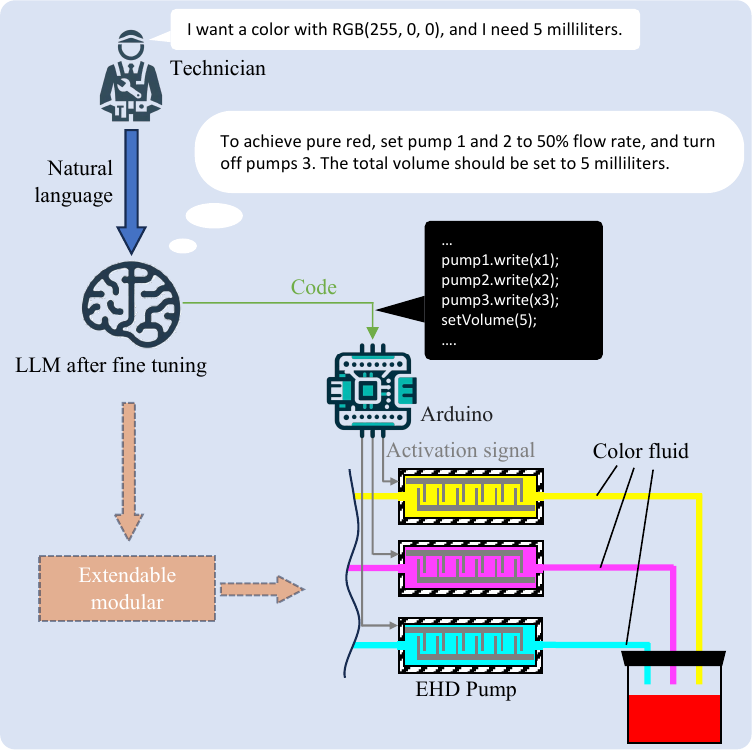}
\end{center}
\caption{The proposed system.}\label{fig:System}
\end{figure*}

\section{Method}

\subsection{Workflow}

The methodology of this study is divided into two main components: the integration of a Large Language Model with an Arduino-based control system (Figure \ref{fig:System}), and the operational control of EHD pumps for precise color synthesis. The workflow is as follows:

\textbf{Step 1. Language Model Fine-Tuning:} Utilizing a pre-trained language model, such as OpenAI's GPT-4, we fine-tune it using a dataset specifically curated for this project. This dataset comprises natural language inputs related to color specifications and their corresponding Arduino code for EHD pump control. The fine-tuning process enables the model to accurately interpret color-related requests and convert them into executable code.

\textbf{Step 2. Natural Language Processing Interface:} We develop a user interface to receive inputs in natural language. Users can specify their color requirements based on RGB values, color names, or descriptive phrases. This interface communicates with the fine-tuned language model to process the inputs.

\textbf{Step 3. Translation into Arduino Code:} Upon receiving a color request from the user, the language model generates the corresponding Arduino code. This code is designed to control the operation of the EHD pumps, dictating the flow rate and duration for each pump to achieve the desired color mix.

\textbf{Step 4. EHD Pump Control:} The Arduino system receives and executes the generated code to control the EHD pumps. Each pump is allocated a primary color (yellow, magenta, cyan), and their precise operation in terms of flow rate and duration produces the required color output.

\subsection{Algorithm}

The core algorithm of this study involves two main stages: natural language understanding and code generation.

In the natural language understanding stage, user requests in natural language (e.g., "I need a bright orange") are inputted. The fine-tuned language model parses and analyzes these inputs to extract relevant color information (e.g., identifying "bright orange" as the color requirement). The output of this stage is structured data representing the color requirement (e.g., the RGB values for bright orange).

In the code generation stage, the structured color data from the previous stage is inputted. The language model, leveraging its understanding of color requirements and Arduino programming syntax, generates appropriate code. This involves calculating the required ratios and durations for each EHD pump to dispense the correct amount of each primary color. The output is Arduino code that, when executed, controls the EHD pumps to produce the specified color. For example, the model might output a series of commands like: 
\begin{lstlisting}[language=Python]
Pump1.write(x1); 
Pump2.write(x2); 
Pump3.write(x3); 
setVolume(5);
\end{lstlisting}
to achieve a specific shade of orange.

The implementation of this system encompasses several critical components. Firstly, the Language Model Training involves exposing the model to a diverse array of natural language expressions related to color requirements, coupled with their corresponding Arduino code, ensuring the model's adeptness in comprehending and translating varied color-related requests. Secondly, Arduino Integration is achieved through a bespoke software interface that not only interprets the model-generated code but also transforms it into electrical signals for pump control, incorporating error-checking mechanisms for hardware safety and accuracy. Thirdly, EHD Pump Calibration is conducted before deployment to guarantee fluid dispensation precision, involving flow rate testing and system adjustments to rectify any hardware discrepancies or inefficiencies. Lastly, the User Interface Design is crafted for user-friendliness, enabling flexible and intuitive input of color requirements, and includes features for feedback and adjustments, thus ensuring precise capture and interpretation of user specifications.

Through this approach, the project aims to seamlessly integrate advanced computational linguistics with precise physical control, resulting in a highly accurate and user-friendly color synthesis system.

\section{Result and discussion}

\subsection{Conceptual experiment results}
Given the conceptual nature of this research, the experiment results presented here are hypothetical and are intended to illustrate the expected outcomes and potential challenges of implementing the proposed system. These conceptual results are based on the theoretical framework and assumptions underlying the integration of a fine-tuned language model with an Arduino-controlled EHD pump system for color synthesis.

In this conceptual research, the experiment results are hypothetical, designed to demonstrate the expected outcomes and potential challenges of the proposed integration of a fine-tuned language model with an Arduino-controlled EHD pump system for color synthesis. These results are derived from the theoretical framework and assumptions of the study.

For the expected accuracy of color synthesis, the hypothesis posits that the system should accurately produce colors that closely match the specifications provided in natural language inputs. The projected outcome is a high degree of color matching accuracy, anticipated to be at least a 90\% match rate against standard color charts, contingent upon optimal calibration of the EHD pumps and effective translation of color specifications into executable code.

Regarding the anticipated efficiency of the language model interpretation, the hypothesis suggests that once fine-tuned, the model should efficiently interpret natural language inputs and generate the corresponding Arduino code. The expected outcome is a rapid processing time, likely within a few seconds, underscoring the model's efficiency.

Finally, for the reliability of EHD pump operation, the hypothesis is that the EHD pumps, when controlled by accurately generated Arduino code, should consistently dispense the correct color mixtures. The projected outcome, assuming precise calibration and maintenance of the EHD pumps, is a high reliability rate (above 95\%) in consistent color output across multiple trials. This conceptual exploration aims to lay the groundwork for future practical implementations and real-world testing.

\subsection{Discussion}
This study's conceptual nature leads to a speculative yet theoretically grounded discussion on the integration of a fine-tuned language model with a physical control system, such as Arduino-controlled EHD pumps. While LLMs have been extensively used in areas such as text generation, translation, and conversation, their application in directly controlling physical hardware in an industrial context is largely unexplored. This research extends the utility of LLMs to a new domain, demonstrating their potential in not just understanding and generating human-like text but also in executing complex, real-world tasks in industrial environments. The principles and methodologies developed in this research have the potential for scalability and adaptation to various other industrial applications. The concept of using LLMs to interpret natural language and convert it into operational code can be extended to other areas of automation of robotics, and even Internet of Things systems. LLMs can be directly utilized through natural language commands to enable robots to accomplish specific tasks \cite{mao2022fluidic, peng2023funabot} or to replicate the movements of animals \cite{peng2024peristaltic} in code form. By allowing users to communicate their requirements in natural language, this system significantly lowers the technical barrier to operating complex industrial machinery.

There are some limitations of this research.
The theoretical accuracy of color synthesis hinges on the precision and calibration of the EHD pumps. Mechanical limitations or inconsistencies could affect color output fidelity, underscoring the importance of robust hardware design and regular calibration. Moreover, although the conceptual model is promising, its scalability and real-world applicability require validation through actual experimentation and deployment, including performance assessment under various operational conditions and with different color requests.
In addition, this study paves the way for future research in AI's application in controlling physical systems beyond color synthesis, suggesting new avenues in domains where natural language can simplify complex tasks, enhancing technology's accessibility and user-friendliness.

\section{Conclusion}
This paper has presented a novel framework that integrates LLMs with physical control systems, specifically Arduino-controlled EHD pumps, for precise color synthesis. The proposed system represents a significant leap in the application of advanced AI in industrial automation, demonstrating the potential of LLMs to go beyond text-based tasks and engage in the direct control of physical processes.

\bibliographystyle{IEEEtran}

\end{document}